# Advancements in Artificial Intelligence Applications for Cardiovascular Disease Research

Yuanlin Mo[1†], Haishan Huang[2†], Bocheng Liang[3], Weibo Ma[4*]

[1] School of Health Science and Engineering, University of Shanghai for Science and Technology, Shanghai, 200093, China
[2] School of Software Engineering, Sun Yat-sen University, Zhuhai, 519000, China
[3] Shenzhen Maternity and Child Healthcare Hospital, Southern Medical University, Shenzhen, 510515, China
[4] School of Nursing and Health Sciences, Shanghai University of Medicine & Health Sciences, Shanghai, 201318, China

*Corresponding author: mwb1030@sina.com;
Contributing authors: moyuanlin g@163.com;
huanghsh25@mail2.sysu.edu.cn;
liangbocheng0519@163.com;
†These authors contributed equally to this work.

**Abstract.** Recent advancements in artificial intelligence (AI) have revolutionized cardiovascular medicine, particularly through integration with computed tomography (CT), magnetic resonance imaging (MRI), electrocardiography (ECG) and ultrasound (US). Deep learning architectures, including convolutional neural networks and generative adversarial networks, enable automated analysis of medical imaging and physiological signals, surpassing human capabilities in diagnostic accuracy and workflow efficiency. However, critical challenges persist, including the inability to validate input data accuracy, which may propagate diagnostic errors. This review highlights AI's transformative potential in precision diagnostics while underscoring the need for robust validation protocols to ensure clinical reliability. Future directions emphasize hybrid models integrating multimodal data and adaptive algorithms to refine personalized cardiovascular care.

**Keywords:** Artificial Intelligence, Cardiovascular Disease, Deep Learning, Cardiac imaging

## 1 Introduction

Recent breakthroughs in artificial intelligence (AI) are profoundly transforming medical practice, particularly cardiovascular medicine, where technological innovations demonstrate unprecedented clinical potential [1]. Contemporary research reveals that advanced computational methodologies that encompass convolutional neural networks (CNNs), Recurrent neural networks (RNNs) and Generative adversarial networks (GANs) are redefining diagnostic paradigms through multimodal analysis of medical images. These advancements not only significantly enhance diagnostic accuracy via pattern recognition that exceeds human capabilities, but also optimize workflow efficiency by automating complex analytical tasks traditionally reliant on expert knowledge. Although previous literature reviews have comprehensively explored the integration of AI with medical sciences, particularly highlighting advancements in visual foundation models for medical image analysis [2], there remains a paucity of comprehensive discussions regarding concrete clinical implementation scenarios.

The integration of AI with advanced medical imaging modalities has significantly enhanced diagnostic capabilities and operational efficiency in cardiovascular healthcare. Computed tomography (CT) technology, characterized by improved resolution, reduced radiation exposure, and shortened scan times, has become indispensable for population health screenings, with AI-driven image analysis further alleviating clinical workloads and lowering healthcare expenditures. Concurrently, magnetic resonance imaging (MRI) retains superiority in soft tissue visualization, particularly for cardiovascular applications, where AI integration not only optimizes reconstruction processes but also augments pathological lesion detection and prognostic prediction accuracy. As the foundational diagnostic tool in cardiology, electrocardiography (ECG) maintains its clinical relevance through cost-effectiveness and radiation-free operation [3], with AI-enhanced interpretation systems emerging as critical solutions to meet growing healthcare demands while complementing advanced imaging modalities. This technological synergy underscores a paradigm shift toward precision medicine through computational augmentation of conventional diagnostic frameworks. Ultrasound (US) offers significant advantages in visualizing cardiovascular structures due to its radiation-free nature, safety, and convenience compared to CT and MRI, making it indispensable for screening fetal cardiovascular diseases. The integration of AI into ultrasound technology holds substantial potential to enhance its diagnostic capabilities.

This review will begin with a brief introduction to the development of deep learning, then systematically examines research advancements across four key domains: 1) AI integration with CT; 2) AI integration with MRI; 3) AI integration with ECG; 4) AI integration with US. By comprehensively evaluating the current research landscape of AI in cardiovascular disease management, our objective is to elucidate existing technological limitations and explore novel developmental directions. This scholarly exploration seeks to provide technical support for achieving precision diagnostics and personalized therapeutic strategies in cardiovascular care.

## 2    The Development of Deep Learning

Deep learning, a transformative subset of artificial intelligence, excels in automated feature extraction and large-scale data processing, enabling breakthroughs in image recognition, signal analysis, and medical diagnostics [4]. Its architecture revolves around multilayered neural networks comprising interconnected input, hidden, and output layers, where weighted connections propagate information hierarchically. The training paradigm combines forward propagation for error computation and back-propagation for iterative weight optimization, refining predictive accuracy through gradient-based adjustments [5].

Prominent deep learning frameworks address diverse data modalities. Convolutional neural networks (CNNs) exploit spatial hierarchies via convolutional and pooling operations, dominating image analysis [6]; Recurrent neural networks (RNNs) process sequential data using cyclic memory [7]; Generative adversarial networks (GANs) synthesize realistic data via adversarial generator-discriminator dynamics, revolutionizing creative applications like image generation and style transfer [8]. Transformer originally introduced in the domain of natural language processing (NLP), has progressively demonstrated its versatility through expanding applications in medical image segmentation [9]. Each architecture exemplifies deep learning's adaptability to heterogeneous problem domains.

## 3 Application of AI in cardiovascular disease

### 3.1 Advancements in AI-Enhanced CT for Cardiovascular Disease

The resolution of computed tomography (CT) images has significantly improved With the advent of spiral CT technology, while the radiation dose to patients has been reduced, and examination times have been substantially shortened. The rapidity and low radiation characteristics of CT technology have led to its widespread application in health screenings. The integration of AI with CT image analysis may not only alleviate the workload of medical staff but also reduces healthcare costs. For instance, Xu et al. [10] proposed a deep learning-based model for the classification of congenital heart diseases. Traditional evaluation using cardiac computed tomography requires manual identification of calcified lesions by operators, whereas Ihdayhid et al. [11] combined AI technology to develop a model capable of automatically obtaining coronary artery calcium (CAC) scoring images. The model classified 1646 (89%) into the same risk category as human observers. Nurmohamed et al. [12] developed a quantitative coronary CT angiography model for diagnosing coronary artery ischemia. Additionally, a model proposed an automated segment-level CAC scoring method for non-contrast CT, enabling precise localization and quantification of calcifications in the coronary artery [13]. Based on AI-driven CT image preprocessing, a model for predicting cardiovascular disease risk was also proposed [14]. This model integrates CAC scores and utilizes segmentation of cardiac chamber size to predict an individual's risk of cardiovascular disease [15], thereby assisting clinicians in evaluating patient conditions and formulating subsequent treatment plans. The specific details of these models are comprehensively presented in Table 1.

**Table 1.** Selected Studies in AI-Enhanced CT for Cardiovascular Disease

| First Author | Purpose | Dataset Size | Strengths | Limitations | Case Type |
|---|---|---|---|---|---|
| Xu et al. | Develop a model for CHD diagnosis | 3,750 | • Higher sensitivity than junior radiologists | • Single-source data<br>• Different CHDs cause bias | CT-1 |
| Ihdayhid et al. | Fully automatedly identifies and quantifies CAC | 5,059 | • Superior accuracy<br>• Rapid processing capability | • Incorrectly reports in zero CAC score | CT-2 |
| Nurmohamed et al. | Diagnose coronary ischemia by invasive FFR and provides prognostic value | 820 | • Superior diagnostic accuracy<br>• Multi-parametric integration | • Limited sample size<br>• Lower performance in CREDENCE | CT-3 |
| Föllmer et al. | Automated segment-level CAC scoring | 1,514 | • Superior accuracy<br>• Multi-task learning | • Lack of consensus between two observers. | CT-4 |
| Miller et al. | Predicting mortality from cardiac volumes mass and coronary calcium | 29,687 | • Better risk classification<br>• Time-efficient workflow | • Lack of specific types of cardiovascular abnormalities | CT-5 |

### 3.2 Advancements in AI-Enhanced MRI for Cardiovascular Disease

Magnetic resonance imaging (MRI) has consistently demonstrated superior capabilities in visualizing soft tissues compared to most other imaging modalities, making it exceptionally effective for cardiovascular imaging. The integration of AI with MRI imaging not only reduces reconstruction time but also enhances the ability of clinicians to identify pathological lesions and predict patient outcomes. For instance, a model is developed for motion-compensated isotropic 3D coronary magnetic resonance angiography, enabling free-breathing acquisition in under a minute[16]. Wang et al. [17] compared the accuracy of three fully automated deep learning algorithms in assessing left ventricular ejection fraction (LVEF), right ventricular ejection fraction (RVEF), and left ventricular mass (LVM), demonstrating that the model's LVEF and LVM measurements were highly consistent with expert results and could be used for clinical decision support. The results show LVEF classification agreed with CLIN-LVEF classification in 86%, 80%, and 85% cases for the 3 DL-LVEF approaches. Additionally, Pezel et al. [18] highlighted that the fully automated AI-derived left atrioventricular coupling index (LACI) provides incremental prognostic value for predicting heart failure. Furthermore, a model proposed the first end-to-end AI system capable of performing full-process CMR interpretation, from screening to multi-disease diagnosis, validating the superiority of video Transformers in medical image analysis. It also revealed that AI can identify CMR features not discernible by human observers. It achieved a match performance with physicians with more than 10 years of experience in CMR reading (F1 score of 0.931 versus 0.927) [19]. The specific details of these models are comprehensively presented in Table 2.

Table 2. Selected Studies in AI-Enhanced MRI for Cardiovascular Disease

| First Author | Purpose | Dataset Size | Strengths | Limitations | Case Type |
|---|---|---|---|---|---|
| Küstner et al. | A framework allows free-breathing acquisitions in a short time | 66 | • Significant improved vessel sharpness | • Only operates on the 2D phase-encoding directions | MRI-1 |
| Wang et al. | Determine whether LVEF, RVEF and LVM measurements can classify ventricular function | 200 | • Good associations with major adverse cardiovascular events | • Low rates of accurate classifications in LVEF | MRI-2 |
| Pezel et al. | Determine whether fully automated AI-based LACI can provide incremental prognostic value to predict HF | 2,134 | • Good accuracy<br>• Improve reclassification beyond traditional HF risk factors | • 8.7% patients were lost to follow-up<br>• Data for medications were not collected | MRI-3 |
| Wang et al. | Demonstrate end-to-end video-models to classify distinct CVDs | 9,719 | • High performance in detect previously unidentified CMR features | • Lower F1 scores for myocarditis<br>• Disability to distinguish phenocopies | MRI-4 |

### 3.3 Advancements in AI-Enhanced ECG for Cardiovascular Disease

As the cornerstone of cardiovascular diagnostics, electrocardiography (ECG) maintains unparalleled clinical utility [3]. Its cost-efficiency and radiation-free nature position it advantageously against advanced imaging modalities like cardiac CT/MRI. This operational reality underscores the critical need for AI-enhanced ECG interpretation systems to address escalating healthcare demands. Contemporary developments demonstrate deep learning's potential across multiple diagnostic domains. Deep learning-based AI models have been developed for automating the interpretation of 12-lead recordings [20], detecting hypertrophic cardiomyopathy signatures [21], identifying paroxysmal atrial fibrillation patterns [22], optimizing rhythm classification accuracy [23], and screening for subclinical systolic dysfunction [24]. The specific details of these models are comprehensively presented in Table 3.

The aforementioned models primarily focus on diagnosing cardiovascular diseases. With the advancement of these models, ECG can also be utilized to more effectively predict the risk of developing cardiovascular diseases. The Pooled Cohort Equations (PCE) remain the clinical gold standard for cardiovascular risk stratification, incorporating quantifiable variables including systolic blood pressure, age, and lipid profiles [26]. Persistent methodological constraints nevertheless undermine their clinical utility. The framework's exclusive reliance on structured numerical inputs neglects critical non-quantifiable variables like psychosocial stressors and dietary patterns. Furthermore, while polygenic risk scores offer hereditary predisposition insights [27], they fail to capture the temporal dynamics of environmental exposures and lifestyle modifications [28], fundamentally limiting their prognostic value. Based on the recognition and diagnosis of electrocardiograms, the SEER system has been developed to assess the risk of cardiovascular diseases. Contemporary prevention paradigms emphasize dynamic risk assessment tools to complement traditional approaches [25].

Table 3. Selected Studies in AI-Enhanced ECG for Cardiovascular Disease

| First Author | Purpose | Dataset Size | Strengths | Limitations | Case Type |
|---|---|---|---|---|---|
| Ribeiro et al. | Develop an end-to-end DNN capable of accurately recognizing six ECG abnormalities | 1,677,211 | • Superior accuracy<br>• Large dataset | • Lack of other classes of abnormalities | ECG-1 |
| Ko et al. | Develop an model for the detection of HCM based on 12-lead ECG | 67,001 | • Superior accuracy<br>• Performed well in younger patients | • Lack of explainability<br>• Exist false positives | ECG-2 |
| Attia et al. | Develop a rapid means of identifying patients with atrial fibrillation | 180,922 | • Superior accuracy<br>• Find signals that might be invisible to the human eye | • Need further prospective calibration inhealthy population | ECG-3 |
| Hughes et al. | Predict long-term risk of cardiovascular mortality and other cardiovascular diseases | 312,422 | • Superior accuracy<br>• Value as a Clinical Supplement | • Demographic biases based on the specific training population | ECG-4 |
| Hong et al. | ECG signal classifcation via combining hand-engineered features | 23,036 | • Superior accuracy<br>• Combine of hand-engineered features and automatic features | • Only study the classification of 7 arrhythmia types | ECG-5 |
| Attia et al. | Permits the ECG to serve as a tool to screen for left ventricular dysfunction | 97,829 | • Superior accuracy<br>• High screening ratein patients without ventricular dysfunction | • ECG–TTE pairs were not simultaneously acquired | ECG-6 |

### 3.4 Advancements in AI-Enhanced US for Cardiovascular Disease

Ultrasound (US) demonstrates significant advantages in visualizing cardiovascular structures. Owing to its radiation-free nature, safety, and convenience compared to CT MRI and US holds an irreplaceable position in screening for fetal cardiovascular diseases. Consequently, the application of AI in US presents substantial potential. Pu et al. pioneered AI applications in cardiac standard plane recognition [29], subsequently developing MobileUNet-FPN for segmenting four chambers of the fetal heart [30], and establishing a new benchmark for unsupervised domain adaptation in fetal cardiac analysis [31]. He et al. implemented AI for detecting standard fetal cardiac planes [32], while Li et al. and He et al. extended its application to fetal cardiac structure detection [33, 34]. Lu et al. and Liang et al. contributed anatomical segmentation algorithms for fetal cardiac structures [35, 36]. Tseng et al. applied AI techniques to M-mode echocardiography analysis [37, 38], while Lu's team introduced SKGC for congenital heart disease classification [39]. Beyond static imaging, Pu et al. further extended AI applications to US video processing for fetal standard cardiac cycle detection [40, 41]. Yang et al. developed video-based learning frameworks for reconstructing cardiac abnormalities from echocardiographic videos [42], and Ouyang et al. explored temporal US video analysis for cardiac function assessment. Trained on echocardiogram videos, This model accurately segments the left ventricle with a Dice similarity coefcient of 0.92, and reliably classifes heart failure with reduced ejection fraction (area under the curve of 0.97) [43]. The specific details of these models are comprehensively presented in Table 4.

**Table 4.** Selected Studies in AI-Enhanced US for Cardiovascular Disease

| First Author | Purpose | Dataset Size | Strengths | Limitations | Case Type |
| --- | --- | --- | --- | --- | --- |
| Liang et al. | Automated measurement of critical anatomical structures in fetal four-chamber view images | 1,083 | • Automated measurement<br>• High segmentation accuracy | • Low performance in RIB<br>• Limited sample size | US-1 |
| Lu et al. | Propose a general semantic-level knowledge-guided classification framework for recognition of fetal CHD | 1,950 | • Good performance with small number of labeled masks | • Limitations of generalization ability | US-2 |
| Tseng et al. | Propose a real-time automatic M-mode echocardiography measurement scheme | 2,639 | • Lossless operation in pixel-unshuffle preserving spatial information<br>• Enabling efficient global attention. | • Inoperable in a 24G GPU due to complexity | US-3 |
| Yang et al. | Proposed CardiacNet for learning morphological abnormalities and motion dysfunction of cardiac disease | 11,527 | • Effectively capture the local structural abnormalities and motion changes | • High calculation cost | US-4 |
| He et al. | Provide efficient technical support for the early detection and diagnosis of fetal cardiac diseases | 29,218 | • High performance of image classification | • Single types of ultrasound images | US-5 |
| Pu et al. | Explores the unsupervised domain adaptive fetal cardiac structure detection issue | 2,879 | • Address pixel-level transformations and topological structure alignment in medical images. | • High Computational Complexity | US-6 |
| Ouyang et al. | Identify subtle changes in ejection fraction lay the foundation for precise diagnosis of cardiovascular disease in real time | 12,925 | • High accuracy<br>• Rapid predictions | • Datasets in limited medical center | US-7 |

## 4 Summary of Selected Studies' Datasets

We summarized the datasets utilized for CT, MRI, ECG and US analyses and presented the statistical results in Figure 1. Given the variability in the number of datasets selected across different models, we calculated the total count of datasets. Additionally, as the total sample sizes across the datasets varied substantially, a logarithmic transformation was applied to the data to better visualize the variations between datasets. As illustrated in Figure 1, ECG demonstrates a significantly larger number of datasets compared to CT, MRI and US. This discrepancy can be attributed to ECG's advantages in easier data acquisition, shorter examination duration, and higher clinical utilization frequency.

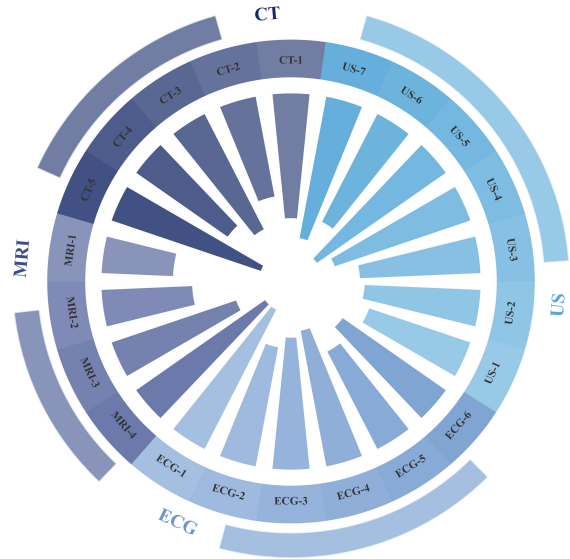

**Fig. 1.** Summary of Selected Studies' Datasets.

## 5   Limitations

One of the challenges in the integration of AI with healthcare is that most AI recognition models cannot verify the correctness of the input data. When there are biases or errors in the input information, the model generates incorrect results, which is particularly concerning in medical applications due to the potential for serious consequences. For example, as shown in Figure 2, Image (a) represents the correct input, and Image (b) is the output result. Image (c) shows incorrect input and Image (d) is the output based on the erroneous input. As the volume of data processed increases, the likelihood of operator errors also rises, making it more difficult to detect incorrect results. This can compromise the authenticity of research outcomes or the accuracy of patient diagnoses and treatment. Therefore, developing a model capable of assessing the correctness of input data is crucial. Currently, the Biomedparse model effectively addresses this issue by accurately distinguishing the correctness of input data [44].

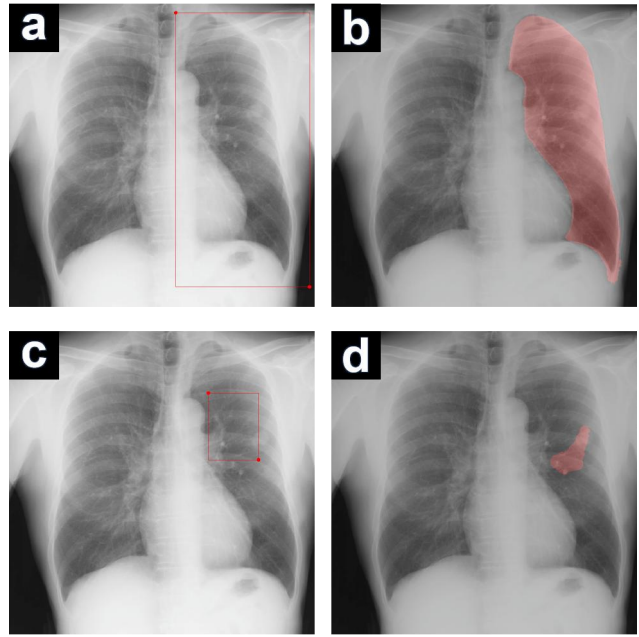

**Fig. 2.** Illustration of AI's inability to recognize input errors[46][47].

The rapid evolution of AI models has yielded increasingly sophisticated architectures with enhanced predictive accuracy. However, this progress has concurrently resulted in escalating computational complexity, posing significant challenges for deploying such models on resource-constrained clinical devices. To address these limitations, research communities have increasingly focused on developing lightweight model frameworks while maintaining diagnostic reliability. Notably, knowledge distillation techniques have emerged as a promising paradigm to enhance the segmentation precision of compact models through hierarchical feature transfer from computationally intensive teacher models [45], thereby achieving optimal balance between inference efficiency and clinical performance. This methodological shift reflects a critical endeavor to bridge the gap between theoretical model advancements and practical clinical applicability in medical imaging workflows.

## 6       Conclusion

The integration of artificial intelligence into cardiovascular medicine is driving transformative advancements in diagnostic and therapeutic paradigms. Deep learning algorithms demonstrate exceptional capabilities across four pivotal domains. Nevertheless, critical challenges persist.Clinical AI deployment stems from models ' inability to validate input data accuracy. Future research must prioritize the development of self-validating AI systems and cross-modal learning frameworks to optimize end-to-end cardiovascular care. Emerging validation frameworks exemplify technological pathways to ensure AI decision reliability, heralding a new era of data-driven medicine synergized with human expertise in cardiovascular healthcare.

## References

1. Patil, S., Shankar, H.: Transforming healthcare: harnessing the power of ai in the modern era. International Journal of Multidisciplinary Sciences and Arts 2(2), 60–70 (2023)
2. Liang, P., Pu, B., Huang, H., Li, Y., Wang, H., Ma, W., Chang, Q.: Vision foundation models in medical image analysis: Advances and challenges. arXiv preprint arXiv:2502.14584 (2025)
3. Hongo, R.H., Goldschlager, N.: Status of computerized electrocardiography. Cardiology clinics 24(3), 491–504 (2006)
4. Shen, D., Wu, G., Suk, H.-I.: Deep learning in medical image analysis. Annual review of biomedical engineering 19(1), 221–248 (2017)
5. Mienye, I.D., Swart, T.G.: A comprehensive review of deep learning: Architectures, recent advances, and applications. Information 15(12), 755 (2024)
6. LeCun, Y., Bottou, L., Bengio, Y., Haffner, P.: Gradient-based learning applied to document recognition. Proceedings of the IEEE 86(11), 2278–2324 (1998)
7. Elman, J.L.: Finding structure in time. Cognitive science 14(2), 179–211 (1990)
8. Goodfellow, I., Pouget-Abadie, J., Mirza, M., Xu, B., Warde-Farley, D., Ozair, S., Courville, A., Bengio, Y.: Generative adversarial nets. Advances in neural information processing systems 27 (2014)
9. Liang, P., Chen, J., Yao, L., Yu, Y., Liang, K., Chang, Q.: Dawtran: dynamic adaptive windowing transformer network for pneumothorax segmentation with implicit feature alignment. Physics in Medicine & Biology 68(17), 175020 (2023)
10. Xu, X., Jia, Q., Yuan, H., Qiu, H., Dong, Y., Xie, W., Yao, Z., Zhang, J., Nie, Z., Li, X., et al.: A clinically applicable ai system for diagnosis of congenital heart diseases based on computed tomography images. Medical Image Analysis 90, 102953 (2023)
11. Ihdayhid, A.R., Lan, N.S., Williams, M., Newby, D., Flack, J., Kwok, S., Joyner, J., Gera, S., Dembo, L., Adler, B., et al.: Evaluation of an artificial intelligence coronary artery calcium scoring model from computed tomography. European radiology 33(1), 321–329 (2023)
12. Nurmohamed, N.S., Danad, I., Jukema, R.A., Winter, R.W., Groot, R.J., Driessen, R.S., Bom, M.J., Diemen, P., Pontone, G., Andreini, D., et al.: Development and validation of a quantitative coronary ct angiography model for diagnosis of vessel-specific coronary ischemia. Cardiovascular Imaging 17(8), 894–906 (2024)
13. Föllmer, B., Tsogias, S., Biavati, F., Schulze, K., Bosserdt, M., Hövermann, L.G., Stober, S., Samek, W., Kofoed, K.F., Maurovich-Horvat, P., et al.: Automated segment-level coronary artery calcium scoring on non-contrast ct: a multi-task deep-learning approach. Insights into Imaging 15(1), 1–12 (2024)
14. Miller, R.J., Killekar, A., Shanbhag, A., Bednarski, B., Michalowska, A.M., Ruddy, T.D., Einstein, A.J., Newby, D.E., Lemley, M., Pieszko, K., et al.: Predicting mortality from ai cardiac volumes mass and coronary calcium on chest computed tomography. Nature Communications 15(1), 2747 (2024)

15. Wasserthal, J., Breit, H.-C., Meyer, M.T., Pradella, M., Hinck, D., Sauter, A.W., Heye, T., Boll, D.T., Cyriac, J., Yang, S., et al.: Totalsegmentator: robust segmentation of 104 anatomic structures in ct images. Radiology: Artificial Intelligence 5(5), 230024 (2023)
16. Küstner, T., Munoz, C., Psenicny, A., Bustin, A., Fuin, N., Qi, H., Neji, R., Kunze, K., Hajhosseiny, R., Prieto, C., et al.: Deep-learning based super-resolution for 3d isotropic coronary mr angiography in less than a minute. Magnetic resonance in medicine 86(5), 2837–2852 (2021)
17. Wang, S., Patel, H., Miller, T., Ameyaw, K., Narang, A., Chauhan, D., Anand, S., Anyanwu, E., Besser, S.A., Kawaji, K., et al.: Ai based cmr assessment of biventricular function: clinical significance of intervendor variability and measurement errors. Cardiovascular Imaging 15(3), 413–427 (2022)
18. Pezel, T., Garot, P., Toupin, S., Sanguineti, F., Hovasse, T., Unterseeh, T., Champagne, S., Morisset, S., Chitiboi, T., Jacob, A.J., et al.: Ai-based fully automated left atrioventricular coupling index as a prognostic marker in patients undergoing stress cmr. Cardiovascular Imaging 16(10), 1288–1302 (2023)
19. Wang, Y.-R., Yang, K., Wen, Y., Wang, P., Hu, Y., Lai, Y., Wang, Y., Zhao, K., Tang, S., Zhang, A., et al.: Screening and diagnosis of cardiovascular disease using artificial intelligence-enabled cardiac magnetic resonance imaging. Nature Medicine 30(5), 1471–1480 (2024)
20. Ribeiro, A.H., Ribeiro, M.H., Paixão, G.M., Oliveira, D.M., Gomes, P.R., Canazart, J.A., Ferreira, M.P., Andersson, C.R., Macfarlane, P.W., Meira Jr, W., et al.: Automatic diagnosis of the 12-lead ecg using a deep neural network. Nature communications 11(1), 1760 (2020)
21. Ko, W.-Y., Siontis, K.C., Attia, Z.I., Carter, R.E., Kapa, S., Ommen, S.R., Demuth, S.J., Ackerman, M.J., Gersh, B.J., Arruda-Olson, A.M., et al.: Detection of hypertrophic cardiomyopathy using a convolutional neural network-enabled electrocardiogram. Journal of the American College of Cardiology 75(7), 722–733 (2020)
22. Attia, Z.I., Noseworthy, P.A., Lopez-Jimenez, F., Asirvatham, S.J., Deshmukh, A.J., Gersh, B.J., Carter, R.E., Yao, X., Rabinstein, A.A., Erickson, B.J., et al.: An artificial intelligence-enabled ecg algorithm for the identification of patients with atrial fibrillation during sinus rhythm: a retrospective analysis of outcome prediction. The Lancet 394(10201), 861–867 (2019)
23. Hong, S., Wu, M., Zhou, Y., Wang, Q., Shang, J., Li, H., Xie, J.: Encase: An ensemble classifier for ecg classification using expert features and deep neural networks. In: 2017 Computing in Cardiology (cinc), pp. 1–4 (2017). IEEE
24. Attia, Z.I., Kapa, S., Lopez-Jimenez, F., McKie, P.M., Ladewig, D.J., Satam, G., Pellikka, P.A., Enriquez-Sarano, M., Noseworthy, P.A., Munger, T.M., et al.: Screening for cardiac contractile dysfunction using an artificial intelligence– enabled electrocardiogram. Nature medicine 25(1), 70–74 (2019)
25. Hughes, J.W., Tooley, J., Torres Soto, J., Ostropolets, A., Poterucha, T., Christensen, M.K., Yuan, N., Ehlert, B., Kaur, D., Kang, G., et al.: A deep learning-based electrocardiogram risk score for long term cardiovascular death and disease. NPJ digital medicine 6(1), 169 (2023)
26. Goff Jr, D.C., Lloyd-Jones, D.M., Bennett, G., Coady, S., D'agostino, R.B., Gibbons, R., Greenland, P., Lackland, D.T., Levy, D., O'donnell, C.J., et al.: 2013 acc/aha guideline on the assessment of cardiovascular risk: a report of the american college of cardiology/american heart association task force on practice guidelines. Circulation 129(25 suppl 2), 49–73 (2014)

27. Riveros-Mckay, F., Weale, M.E., Moore, R., Selzam, S., Krapohl, E., Sivley, R.M., Tarran, W.A., Sørensen, P., Lachapelle, A.S., Griffiths, J.A., et al.: Integrated polygenic tool substantially enhances coronary artery disease prediction. Circulation: Genomic and Precision Medicine 14(2), 003304 (2021)
28. Hughes, J.W., Tooley, J., Torres Soto, J., Ostropolets, A., Poterucha, T., Christensen, M.K., Yuan, N., Ehlert, B., Kaur, D., Kang, G., et al.: A deep learning-based electrocardiogram risk score for long term cardiovascular death and disease. NPJ digital medicine 6(1), 169 (2023)
29. Pu, B., Li, K., Li, S., Zhu, N.: Automatic fetal ultrasound standard plane recognition based on deep learning and iiot. IEEE Transactions on Industrial Informatics 17(11), 7771–7780 (2021)
30. Pu, B., Lu, Y., Chen, J., Li, S., Zhu, N., Wei, W., Li, K.: Mobileunet-fpn: A semantic segmentation model for fetal ultrasound four-chamber segmentation in edge computing environments. IEEE Journal of Biomedical and Health Informatics 26(11), 5540–5550 (2022)
31. Pu, B., Wang, L., Yang, J., He, G., Dong, X., Li, S., Tan, Y., Chen, M., Jin, Z., Li, K., et al.: M3-uda: a new benchmark for unsupervised domain adaptive fetal cardiac structure detection. In: Proceedings of the IEEE/CVF Conference on Computer Vision and Pattern Recognition, pp. 11621–11630 (2024)
32. He, J., Yang, L., Liang, B., Li, S., Xu, C.: Fetal cardiac ultrasound standard section detection model based on multitask learning and mixed attention mechanism. Neurocomputing 579, 127443 (2024)
33. Li, X., Tan, Y., Liang, B., Pu, B., Yang, J., Zhao, L., Kong, Y., Yang, L., Zhang, R., Li, H., et al.: Tkr-fsod: Fetal anatomical structure few-shot detection utilizing topological knowledge reasoning. IEEE Journal of Biomedical and Health Informatics (2024)
34. He, J., Yang, L., Zhu, Y., Li, D., Ding, Z., Lu, Y., Liang, B., Li, S.: Fetal cardiac structure detection using multi-task learning. In: International Conference on Intelligent Computing, pp. 405–419 (2024). Springer
35. Liang, B., Peng, F., Luo, D., Zeng, Q., Wen, H., Zheng, B., Zou, Z., An, L., Wen, H., Wen, X., et al.: Automatic segmentation of 15 critical anatomical labels and measurements of cardiac axis and cardiothoracic ratio in fetal four chambers using nnu-netv2. BMC Medical Informatics and Decision Making 24(1), 128 (2024)
36. Lu, Y., Li, K., Pu, B., Tan, Y., Zhu, N.: A yolox-based deep instance segmentation neural network for cardiac anatomical structures in fetal ultrasound images. IEEE/ACM Transactions on Computational Biology and Bioinformatics (2022)
37. Tseng, C.-H., Chien, S.-J., Wang, P.-S., Lee, S.-J., Pu, B., Zeng, X.-J.: Real-time automatic m-mode echocardiography measurement with panel attention. IEEE Journal of Biomedical and Health Informatics (2024)
38. Tseng, C.-H., Chien, S.-J., Wang, P.-S., Lee, S.-J., Hu, W.-H., Pu, B., Zeng, X.-j.: Real-time automatic m-mode echocardiography measurement with panel attention from local-to-global pixels. arXiv preprint arXiv:2308.07717 (2023)
39. Lu, Y., Tan, G., Pu, B., Wang, H., Liang, B., Li, K., Rajapakse, J.C.: Skgc: A general semantic-level knowledge guided classification framework for fetal congenital heart disease. IEEE Journal of Biomedical and Health Informatics (2024)
40. Pu, B., Zhu, N., Li, K., Li, S.: Fetal cardiac cycle detection in multi-resource echocardiograms using hybrid classification framework. Future Generation Computer Systems 115, 825–836 (2021)

41. Pu, B., Li, K., Chen, J., Lu, Y., Zeng, Q., Yang, J., Li, S.: Hfsccd: a hybrid neural network for fetal standard cardiac cycle detection in ultrasound videos. IEEE Journal of Biomedical and Health Informatics (2024)
42. Yang, J., Lin, Y., Pu, B., Guo, J., Xu, X., Li, X.: Cardiacnet: Learning to reconstruct abnormalities for cardiac disease assessment from echocardiogram videos. In: European Conference on Computer Vision, pp. 293–311 (2024). Springer
43. Ouyang, D., He, B., Ghorbani, A., Yuan, N., Ebinger, J., Langlotz, C.P., Heidenreich, P.A., Harrington, R.A., Liang, D.H., Ashley, E.A., et al.: Video-based ai for beat-to-beat assessment of cardiac function. Nature 580(7802), 252–256 (2020)
44. Zhao, T., Gu, Y., Yang, J., Usuyama, N., Lee, H.H., Naumann, T., Gao, J., Crabtree, A., Abel, J., Moung-Wen, C., et al.: Biomedparse: a biomedical foundation model for image parsing of everything everywhere all at once. arXiv preprint arXiv:2405.12971 (2024)
45. Liang, P., Chen, J., Chang, Q., Yao, L.: Rskd: Enhanced medical image segmentation via multi-layer, rank-sensitive knowledge distillation in vision transformer models. Knowledge-Based Systems 293, 111664 (2024)
46. Shiraishi, J., Katsuragawa, S., Ikezoe, J., Matsumoto, T., Kobayashi, T., Komatsu, K.-i., Matsui, M., Fujita, H., Kodera, Y., Doi, K.: Development of a digital image database for chest radiographs with and without a lung nodule: receiver operating characteristic analysis of radiologists' detection of pulmonary nodules. American journal of roentgenology 174(1), 71–74 (2000)
47. Ma, J., He, Y., Li, F., Han, L., You, C., Wang, B.: Segment anything in medical images. Nature Communications 15(1), 654 (2024)